\pdfoutput=1

\documentclass[11pt]{article}

\usepackage[preprint]{acl}

\usepackage{times}
\usepackage{latexsym}

\usepackage[T1]{fontenc}

\usepackage[utf8]{inputenc}

\usepackage{microtype}

\usepackage{inconsolata}

\usepackage{graphicx}

\usepackage{amsmath,amsfonts}
\usepackage{enumitem}

\usepackage{tikz}
\usepackage[edges]{forest}
\definecolor{hidden-draw}{RGB}{20,68,106}
\definecolor{hidden-pink}{RGB}{255,245,247}

\usepackage{booktabs}
\usepackage{lscape}
\usepackage{array}
\newcolumntype{T}{>{\ttfamily}c}
\usepackage{lipsum}

\usepackage{colortbl}
\usepackage{wrapfig}

\title{A Survey on LLM Inference-Time Self-Improvement}

\author{Xiangjue Dong\thanks{\hspace{0.2cm}Equal Contribution}\quad Maria Teleki\footnotemark[1]\quad James Caverlee \\Texas A\&M University \\ \small\texttt{\{xj.dong,mariateleki,caverlee\}@tamu.edu}}

\begin{document}
\maketitle
\begin{abstract}

Techniques that enhance inference through increased computation at test-time have recently gained attention. 
In this survey, we investigate the current state of LLM Inference-Time Self-Improvement from three different perspectives: Independent Self-improvement, focusing on enhancements via decoding or sampling methods; Context-Aware Self-Improvement, leveraging additional context or datastore; and Model-Aided Self-Improvement, achieving improvement through model collaboration. We provide a comprehensive review of recent relevant studies, contribute an in-depth taxonomy, and discuss challenges and limitations, offering insights for future research. 
\end{abstract}

\noindent
\begin{wrapfigure}{l}{0.05\textwidth}
    \centering
    \hypertarget{github-link}{}
    \href{https://github.com/dongxiangjue/Awesome-LLM-Self-Improvement}{%
    \includegraphics[width=0.05\textwidth]{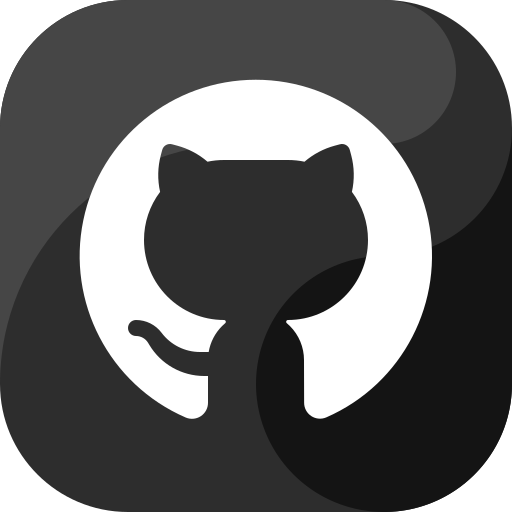}
    }
\vspace{-2em}
\end{wrapfigure}

\noindent
{\url{https://github.com/dongxiangjue/Awesome-LLM-Self-Improvement}}

\section{Introduction} \label{sec:introduction}

The capabilities of large language models (LLMs) have advanced dramatically in recent years~\cite{achiam2023gpt, gemini}.
These advancements have largely been driven by scaling up model training compute~\cite{kaplan2020scalinglawsneurallanguage,brown2024large}, with investments in larger models, extensive pretraining datasets, and enhanced alignment techniques~\cite{ouyang2022training,bai2022training,bai2022constitutional,rafailov2023direct}.\footnote{These investments are expensive in terms of both capital and carbon footprint -- to the point that technology companies are investing in nuclear power \cite{nuclear1}.} 

\begin{figure}
    \centering
    \includegraphics[width=\columnwidth]{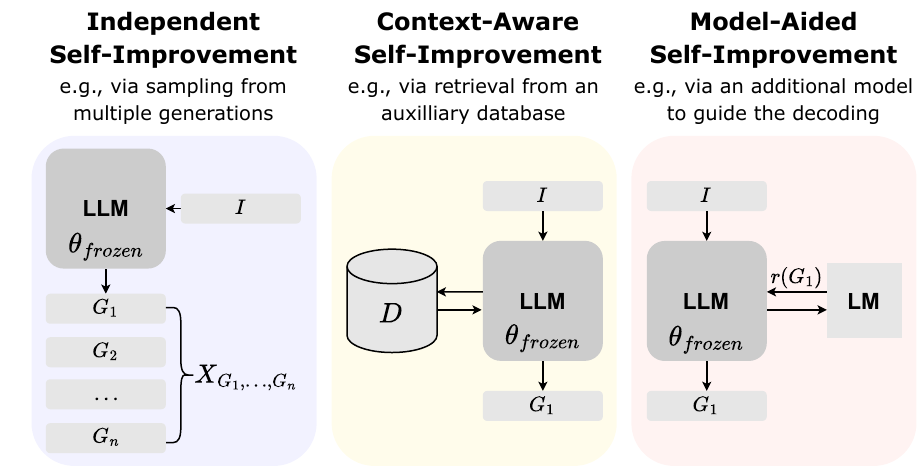}
    \caption{Conceptual examples of methods in the three main categories of \textit{LLM Inference-Time Self-Improvement} without altering the original LLM parameters or additional training.}
    \label{fig:llm}
\end{figure}

\tikzstyle{my-box}=[
    rectangle,
    draw=hidden-draw,
    rounded corners,
    text opacity=1,
    minimum height=1.5em,
    minimum width=5em,
    inner sep=2pt,
    align=center,
    fill opacity=.5,
    line width=0.8pt,
]
\tikzstyle{leaf}=[my-box, minimum height=1.5em,
    fill=blue!10, text=black, align=left,font=\normalsize,
    inner xsep=2pt,
    inner ysep=4pt,
    line width=0.8pt,
]
\tikzstyle{leaf1}=[my-box, minimum height=1.5em,
    fill=yellow!20, text=black, align=left,font=\normalsize,
    inner xsep=2pt,
    inner ysep=4pt,
    line width=0.8pt,
]
\tikzstyle{leaf2}=[my-box, minimum height=1.5em,
    fill=red!10, text=black, align=left,font=\normalsize,
    inner xsep=2pt,
    inner ysep=4pt,
    line width=0.8pt,
]
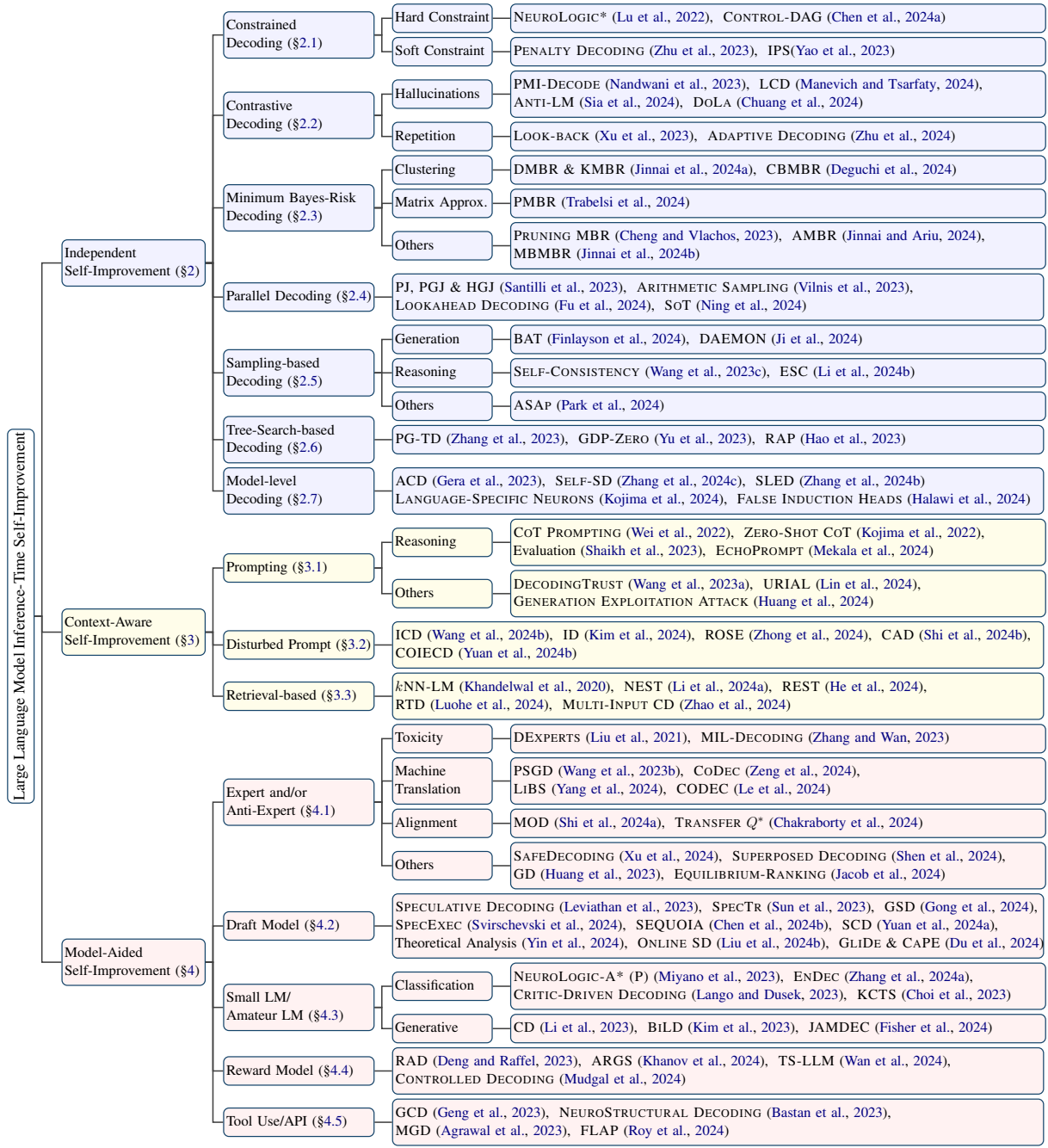
\begin{figure*}[ht!]
    \vspace{-1.0cm}
    \centering
    \resizebox{\textwidth}{!}{
        \begin{forest}
            forked edges,
            for tree={
                grow=east,
                reversed=true,
                anchor=base west,
                parent anchor=east,
                child anchor=west,
                base=left,
                font=\large,
                rectangle,
                draw=hidden-draw,
                rounded corners,
                align=left,
                minimum width=4em,
                edge+={darkgray, line width=1pt},
                s sep=3pt,
                inner xsep=2pt,
                inner ysep=3pt,
                line width=0.8pt,
                ver/.style={rotate=90, child anchor=north, parent anchor=south, anchor=center},
            },
            where level=1{text width=9.5em,font=\normalsize,}{},
            where level=2{text width=10.0em,font=\normalsize}{},
            where level=3{text width=6.5em,font=\normalsize,}{},
            where level=4{text width=6.0em,font=\normalsize,}{},
            [
                Large Language Model Inference-Time Self-Improvement, ver
                [
                    Independent\\Self-Improvement (\S \ref{sec:self})
                    , leaf
                    [
                        Constrained\\Decoding (\S \ref{ssec:constrained})
                        , leaf
                        [
                            Hard Constraint
                            , leaf
                            [
                                \textsc{NeuroLogic*}~\cite{lu-etal-2022-neurologic}{, }
                                \textsc{Control-DAG}~\cite{chen-etal-2024-control}
                                , leaf, text width=36.5em
                            ]
                        ]
                        [
                            Soft Constraint
                            , leaf
                            [
                               \textsc{Penalty Decoding}~\cite{zhu-etal-2023-penalty}{, }
                               \textsc{IPS}\cite{yao-etal-2023-fine}
                                , leaf, text width=36.5em
                            ]
                        ]
                    ]
                    [
                        Contrastive\\Decoding (\S \ref{ssec:contrastive})
                        , leaf
                        [
                            Hallucinations
                            , leaf
                            [
                               \textsc{PMI-Decode}~\cite{nandwani-etal-2023-pointwise}{, }
                               \textsc{LCD}~\cite{manevich-tsarfaty-2024-mitigating}{, }\\
                               \textsc{Anti-LM}~\cite{sia-etal-2024-anti}{, }
                               \textsc{DoLa}~\cite{chuang2024dola}
                                , leaf, text width=36.5em
                            ]
                        ]
                        [
                            Repetition
                            , leaf
                            [
                               \textsc{Look-back}~\cite{xu-etal-2023-look}{, }
                               \textsc{Adaptive Decoding}~\cite{zhu2024improving}
                                , leaf, text width=36.5em
                            ]
                        ]
                    ]
                    [
                        Minimum Bayes-Risk\\Decoding (\S \ref{ssec:MBR})
                        , leaf
                        [
                            Clustering
                            , leaf
                            [
                               \textsc{DMBR} \& \textsc{KMBR}~\cite{jinnai-etal-2024-generating}{, }
                               \textsc{CBMBR}~\cite{deguchi-etal-2024-centroid}
                                , leaf, text width=36.5em
                            ]
                        ]
                        [
                            Matrix Approx.
                            , leaf
                            [
                               \textsc{PMBR}~\cite{trabelsi2024efficient}
                                , leaf, text width=36.5em
                            ]
                        ]
                        [
                            Others
                            , leaf
                            [
                               \textsc{Pruning MBR}~\cite{cheng-vlachos-2023-faster}{, }
                               \textsc{AMBR}~\cite{jinnai-ariu-2024-hyperparameter}{, }\\
                               \textsc{MBMBR}~\cite{jinnai2024modelbased}
                                , leaf, text width=36.5em
                            ]
                        ]
                    ]
                    [
                        Parallel Decoding (\S \ref{ssec:parallel})
                        , leaf
                        [
                            \textsc{PJ}{, }\textsc{PGJ} \& \textsc{HGJ}~\cite{santilli-etal-2023-accelerating}{, }
                            \textsc{Arithmetic Sampling}~\cite{pmlr-v202-vilnis23a}{, }\\
                            \textsc{Lookahead Decoding}~\cite{fu2024break}{, }
                            \textsc{SoT}~\cite{ning2024skeletonofthought}       
                            , leaf, text width=44.5em
                        ]
                    ]
                    [
                        Sampling-based\\Decoding (\S \ref{ssec:sampling})
                        , leaf
                        [
                            Generation
                            , leaf
                            [
                               \textsc{BAT}~\cite{finlayson2024closing}{, }
                               \textsc{DAEMON}~\cite{ji2024language}
                                , leaf, text width=36.5em
                            ]
                        ]
                        [
                            Reasoning
                            , leaf
                            [
                               \textsc{Self-Consistency}~\cite{wang2023selfconsistency}{, }
                               \textsc{ESC}~\cite{li2024escape}
                                , leaf, text width=36.5em
                            ]
                        ]
                        [
                            Others
                            , leaf
                            [
                               \textsc{ASAp}~\cite{park2024grammaraligned}
                                , leaf, text width=36.5em
                            ]
                        ]
                    ]                    
                    [
                        Tree-Search-based\\Decoding (\S \ref{ssec:tree})
                        , leaf
                        [
                            \textsc{PG-TD}~\cite{zhang2023planning}{, }
                            \textsc{GDP-Zero}~\cite{yu-etal-2023-prompt}{, }
                            \textsc{RAP} \cite{hao-etal-2023-reasoning}
                            , leaf, text width=44.5em
                        ]
                    ]
                    [
                        Model-level\\Decoding (\S \ref{ssec:model})
                        , leaf
                        [
                            \textsc{ACD} \cite{gera2023autocontrastive}{, }
                            \textsc{Self-SD}~\cite{zhang-etal-2024-draft}{, }
                            \textsc{SLED}~\cite{zhang2024sled}\\
                            \textsc{Language-Specific Neurons}~\cite{kojima-etal-2024-multilingual}{, }
                            \textsc{False Induction Heads}~\cite{halawi2024overthinking}\\
                            , leaf, text width=44.5em
                        ]
                    ]
                ]    
                [
                    Context-Aware\\Self-Improvement (\S \ref{sec:contextawareselfimprovement})
                    , leaf1
                    [
                        Prompting (\S \ref{ssec:prompting})
                        , leaf1
                        [
                            Reasoning
                            , leaf1
                            [
                               \textsc{CoT Prompting} \cite{wei2022chain}{, }
                               \textsc{Zero-Shot CoT} \cite{kojima2022large}{, }\\
                               Evaluation \cite{shaikh-etal-2023-second}{, }
                               \textsc{EchoPrompt} \cite{mekala-etal-2024-echoprompt}
                               , leaf1, text width=36.5em
                            ]
                        ]
                        [
                            Others
                            , leaf1
                            [
                               \textsc{DecodingTrust} \cite{wang2023decodingtrust}{, }
                               \textsc{URIAL} \cite{lin2024the}{, } \\
                               \textsc{Generation Exploitation Attack} \cite{huang2024catastrophic}
                               , leaf1, text width=36.5em
                            ]
                        ]
                    ]
                    [
                        Disturbed Prompt (\S \ref{ssec:disturbed})
                        , leaf1
                        [  
                           \textsc{ICD} \cite{wang-etal-2024-mitigating}{, }
                           \textsc{ID} \cite{kim2024instructive}{, }
                           \textsc{ROSE} \cite{zhong-etal-2024-rose}{, }
                           \textsc{CAD} \cite{shi-etal-2024-trusting}{, }\\
                           \textsc{COIECD}~\cite{yuan-etal-2024-discerning}
                           ,leaf1, text width=44.5em
                        ]
                    ]
                    [
                        Retrieval-based (\S \ref{ssec:retrievalbased})
                        , leaf1
                        [
                            \textsc{$k$NN-LM} \cite{Khandelwal2020Generalization}{, }
                            \textsc{NEST} \cite{li2024nearest}{, }
                            \textsc{REST} \cite{he-etal-2024-rest}{, }\\
                            \textsc{RTD} \cite{luohe2024reference}{, }
                            \textsc{Multi-Input CD} \cite{zhao-etal-2024-enhancing}
                            ,leaf1, text width=44.5em
                        ]
                    ]
                ]
                [
                    Model-Aided\\Self-Improvement (\S \ref{sec:modelaided})
                    , leaf2
                    [
                        Expert and/or\\Anti-Expert (\S \ref{sec:expert-anti-expert})
                        , leaf2
                        [
                            Toxicity
                            , leaf2
                            [
                               \textsc{DExperts} \cite{liu-etal-2021-dexperts}{, }
                               \textsc{MIL-Decoding} \cite{zhang-wan-2023-mil}
                               , leaf2, text width=36.5em
                            ]
                        ]
                        [
                            Machine\\Translation
                            , leaf2
                            [
                               \textsc{PSGD} \cite{wang-etal-2023-easy}{, }
                               \textsc{CoDec} \cite{zeng-etal-2024-improving}{, }\\
                               \textsc{LiBS} \cite{yang-etal-2024-language-informed}{, }
                               \textsc{CODEC} \cite{le2024constrained}
                               , leaf2, text width=36.5em
                            ]
                        ]
                        [
                            Alignment
                            , leaf2
                            [
                               \textsc{MOD} \cite{shi2024decodingtime}{, }
                               \textsc{Transfer $Q^*$} \cite{chakraborty2024transfer}
                               , leaf2, text width=36.5em
                            ]
                        ]
                        [
                            Others
                            , leaf2
                            [
                               \textsc{SafeDecoding} \cite{xu-etal-2024-safedecoding}{, }
                               \textsc{Superposed Decoding} \cite{shen2024superposed}{, }\\
                               \textsc{GD} \cite{huang2023grounded}{, }
                               \textsc{Equilibrium-Ranking} \cite{jacob2024the}
                               , leaf2, text width=36.5em
                            ]
                        ]
                    ]
                    [
                        Draft Model (\S \ref{sec:draft-model})
                        , leaf2
                        [
                            \textsc{Speculative Decoding} \cite{pmlr-v202-leviathan23a}{, }
                            \textsc{SpecTr} \cite{sun2023spectr}{, }
                            \textsc{GSD} \cite{gong-etal-2024-graph}{, }\\
                            \textsc{SpecExec} \cite{svirschevski2024specexec}{, }
                            \textsc{SEQUOIA} \cite{chen2024sequoia}{, }
                            \textsc{SCD} \cite{yuan-etal-2024-speculative}{, }\\
                            Theoretical Analysis \cite{yin2024a}{, }
                            \textsc{Online SD} \cite{liu2024online}{, }
                            \textsc{GliDe} \& \textsc{CaPE} \cite{du2024glide}
                            , leaf2, text width=44.5em
                        ]
                    ]
                    [
                        Small LM/\\Amateur LM (\S \ref{ssec:smalllm})
                        , leaf2
                        [
                            Classification
                            , leaf2
                            [
                               \textsc{NeuroLogic-A* (P)} \cite{miyano-etal-2023-self}{, }
                               \textsc{EnDec} \cite{zhang-etal-2024-jailbreak}{, }\\
                               \textsc{Critic-Driven Decoding} \cite{lango-dusek-2023-critic}{, }
                               \textsc{KCTS} \cite{choi-etal-2023-kcts}
                               , leaf2, text width=36.5em
                            ]
                        ]
                        [
                            Generative
                            , leaf2
                            [
                               \textsc{CD} \cite{li-etal-2023-contrastive}{, }
                               \textsc{BiLD} \cite{kim2023speculative}{, }
                               \textsc{JAMDEC} \cite{fisher-etal-2024-jamdec}
                               , leaf2, text width=36.5em
                            ]
                        ]
                    ]
                    [
                        Reward Model (\S \ref{sec:reward-model})
                        , leaf2
                        [ 
                            \textsc{RAD} \cite{deng-raffel-2023-reward}{, }
                            \textsc{ARGS} \cite{khanov2024args}{, }
                            \textsc{TS-LLM} \cite{wan2024alphazerolike}{, }\\
                            \textsc{Controlled Decoding} \cite{mudgal2024controlled}
                            , leaf2, text width=44.5em
                        ]
                    ]
                    [
                        Tool Use\//API (\S \ref{sec:tool-use})
                        , leaf2
                        [ 
                            \textsc{GCD} \cite{geng-etal-2023-grammar}{, }
                            \textsc{NeuroStructural Decoding} \cite{bastan-etal-2023-neurostructural}{, }\\
                            \textsc{MGD} \cite{agrawal2023monitorguided}{, }
                            \textsc{FLAP} \cite{roy-etal-2024-flap}
                            , leaf2, text width=44.5em
                        ]
                    ]
                ]  
            ]
        \end{forest}
        }
    \caption{Taxonomy of Large Language Model Inference-Time Self-Improvement.}
    \label{fig:taxonomy}
\end{figure*}

Recently, scaling computation \textit{during} inference-time to improve task performance has gained attention~\cite{snell2024scaling}, e.g., increasing test-time compute (i.e., model thinking time)~\cite{openai_learning_reason} and scaling inference compute through repeated sampling~\cite{brown2024large}. Test-time capabilities enable smaller models to replace larger ones by trading size for extra inference computation and pave the way for self-improvement with minimal human supervision~\cite{brown2024large}. 
Self-improvement approaches at inference-time offer a new set of opportunities for researchers to continue pushing the boundaries of AI models beyond scaling model size and training data. Concurrent reviews about methods at inference-time are from the perspective of (i) token-level generation versus meta-generation \cite{welleck2024decoding}, and (ii) verification methods \cite{song2024mind}. However, there has not yet been a comprehensive review of self-improvement at inference-time. This survey aims to bridge this gap by categorically reviewing existing literature and contributing the first taxonomy (Figure \ref{fig:taxonomy}) of these methods.

We survey in this work this notion of \textbf{LLM Inference-Time Self Improvement} (ITSI\footnote{Pronounced \textit{``itsy.''}}) -- \textit{i.e., relying on the LLM's own frozen parameters without additional training or parameter updating to improve performance and/or efficiency at inference-time.} These methods often incorporate specialized decoding algorithms to refine the next-token selection by adjusting logits (raw model outputs), the probability distribution (\textit{softmax} applied to the logits), and the decoding objective. 
We classify these methods into three categories: \textbf{Independent Self-Improvement}, which operates independently; \textbf{Context-Aware Self-Improvement}, which leverages external support (i.e. context and datastore retrieval); and \textbf{Model-Aided Self-Improvement}, which relies on external models for collaboration. 
We draw from recent high-quality papers published in top conferences (ACL, EMNLP, NAACL, NeurIPS, ICLR, ICML) and highly cited works. 
We highlight key challenges and potential directions for future research. 

\section{Independent Self-Improvement}
\label{sec:self}


Independent Self-Improvement is achieving improvements in performance using the model's own frozen parameters without additional training -- i.e., by modifying the decoding process (\S \ref{ssec:constrained}, \S \ref{ssec:contrastive}, and \S \ref{ssec:MBR}); increasing efficiency (\S \ref{ssec:parallel}); sampling multiple candidate generations (\S \ref{ssec:sampling}); and isolating layers or neurons (\S \ref{ssec:model}).

\subsection{Constrained Decoding}
\label{ssec:constrained}

Constrained decoding guides the generation process via hard constraints or soft constraints.


\smallskip \noindent \textbf{Hard Constraint.}
These are strict rules that must be adhered to in the output, such as requiring a specific word to appear in the generated sentence.
\citet{lu-etal-2022-neurologic} propose \textsc{NeuroLogic*} (NeuroLogic A*esque Decoding), extending NeuroLogic Decoding~\cite{lu-etal-2021-neurologic} with lookahead heuristics to estimate future constraint satisfaction and enforce logical constraints.
\citet{chen-etal-2024-control} propose \textsc{Control-DAG}, a constrained decoding algorithm for
Directed Acyclic models. It enables \textit{lexical, vocabulary, and length controls} to ensure the generation of designated entities (\textit{lexical constraints}), eliminate out-of-vocabulary words (\textit{vocabulary constraints}), and regulate the target length of the output using a Viterbi decoding algorithm (\textit{length constraints}). 

\smallskip \noindent \textbf{Soft Constraint.} 
These are more flexible guidelines that the model aims to satisfy, but they are not strictly enforced. 
The model attempts to generate text that adheres to these constraints while still optimizing for fluency and relevance.
\citet{zhu-etal-2023-penalty} propose \textsc{Penalty Decoding} via a forgetting mechanism -- i.e., an adjustment to the probability distribution with a repetition penalty. 
\citet{yao-etal-2023-fine} propose \textsc{IPS} (Isotropic and Proximal Search), wherein they modify the generated token selection via terms for (i) isotropy -- i.e., responses should be similar to all previous turns, and (ii) proximity -- i.e., generated tokens should be similar to previously generated tokens. These criteria are incorporated via a weighted penalty in the decoding objective.

\subsection{Contrastive Decoding}
\label{ssec:contrastive}

Contrastive decoding adjusts the next-token probability based on differences in logits.

\smallskip \noindent \textbf{Faithfulness and Hallucinations.}
\citet{nandwani-etal-2023-pointwise} propose \textsc{PMI-Decode} (Point-wise Mutual Information Decode), a decoding method which incorporates \textsc{PMI-Faith} in its objective to ensure faithfulness to the text. \textsc{PMI-Faith} is based on conditional (dialog history) pointwise mutual information between the generated response and the document. 
\citet{manevich-tsarfaty-2024-mitigating} propose \textsc{LCD} (Language Contrastive Decoding) which adjusts the output probabilities of a large vision-language model based on its internal LLM's probability distribution using \textit{a dynamic weighting mechanism guided by entropy}. 
\citet{sia-etal-2024-anti} propose \textsc{Anti-LM} (Anti-Language Model), a decoding objective which applies exponential decay to penalize the logits of the next token conditioned on the test sentences being translated -- without considering other contexts or subsequent generations. 
\citet{chuang2024dola} propose \textsc{DoLa} (Decoding by Contrasting Layers) which improves model factuality by contrasting logits from higher and lower layers. 

\smallskip \noindent \textbf{Repetition, Coherence, and Diversity.}
\citet{xu-etal-2023-look} propose \textsc{Look-back} decoding, which relies on KL-Divergence to avoid token/phrase repetition and topic changes.
\citet{zhu2024improving} propose \textsc{Adaptive Decoding} that dynamically adjusts the candidate set size through a confidence-increasing procedure based on entropy. Tokens with the highest probability are selected to enhance confidence, updating the candidate set to determine the optimal set during generation.


\subsection{Minimum Bayes-Risk Decoding}
\label{ssec:MBR}

Unlike MAP decoding, Minimum Bayes-Risk (MBR) decoding selects the output sentence that maximizes the \textit{expected utility} over the set of translation hypotheses~\cite{kumar-byrne-2004-minimum}.

\smallskip \noindent \textbf{Clustering.} 
\citet{jinnai-etal-2024-generating} propose \textsc{DMBR} (Diverse MBR) and \textsc{KMBR} ($k$-medoids MBR): DMBR adds a diversity penalty to optimize utility and diversity, while KMBR reframes decoding as clustering. 
\citet{deguchi-etal-2024-centroid} propose \textsc{CBMBR} (Centroid-Based MBR), which accelerates MBR decoding by clustering reference translations in feature space and employing cluster centroids to compute \textit{expected utility}.

\smallskip \noindent \textbf{Matrix Approximation.} 
\citet{trabelsi2024efficient} propose \textsc{PMBR} (Probabilistic MBR decoding), which leverages the MBR matrix's low-rank structure and applies matrix completion to a subset of scores to estimate the full matrix.

\smallskip \noindent \textbf{Others.} 
\citet{cheng-vlachos-2023-faster} propose \textsc{Pruning MBR}, which speeds up traditional MBR by pruning low-utility hypotheses via sampling estimation. 
Improving on this, \citet{jinnai-ariu-2024-hyperparameter} propose \textsc{AMBR} (Adaptive Minimum Bayes-Risk), which uses the \textit{Correlated Sequential Halving} algorithm to compute the sample-based MBR decoding objective and automatically optimize resource allocation based on the computational budget. 
\citet{jinnai2024modelbased} propose \textsc{MBMBR} (Model-Based MBR), replacing the Monte Carlo estimate with the model's own probability distribution.

\subsection{Parallel Decoding}
\label{ssec:parallel}

Parallel decoding generates multiple tokens simultaneously during the decoding phases for faster generation, rather than sequentially.
For machine translation tasks, \citet{santilli-etal-2023-accelerating} propose \textsc{HGJ} (Hybrid GS-Jacobi Decoding) for ``a parallel formulation leveraging Jacobi and Gauss-Seidel fixed-point iteration methods for fast inference.'' 
\citet{pmlr-v202-vilnis23a} propose \textsc{Arithmetic Sampling}, which generates diverse, unbiased samples from LLMs using an implicitly defined arithmetic code book. 
\citet{fu2024break} propose \textsc{Lookahead Decoding}, a parallel algorithm that accelerates LLM decoding by generating and verifying $n$-grams in a single step.
\citet{ning2024skeletonofthought} propose \textsc{SoT} (Skeleton-of-Thought), which generates an answer outline, fills in details via batched decoding or parallel API calls, and aggregates the results.

\subsection{Sampling-based Decoding}
\label{ssec:sampling}

Sampling-based methods introduce randomness for token selection to generate diverse text or sample multiple generation paths from the model.

\smallskip \noindent \textbf{Open-Ended Generation.} 
\citet{finlayson2024closing} propose \textsc{BAT} (Basis-Aware Truncation) sampling, based on \textit{the softmax bottleneck} \cite{yang2017breaking}. BAT applies two constraints -- a threshold and a basis-aware constraint -- using a linear programming solver. It discards higher-probability tokens while keeping lower-probability, higher-quality tokens, achieving strong results in low-entropy open-ended generation.
\citet{ji2024language} propose \textsc{DAEMON} (Decoding As DirEct Metrics OptimizatioN), which draws on a \textit{``sampling-importance-resampling''} process. It identifies a decoding distribution that aligns sampled texts with human texts by searching for the decoding distribution that minimizes the reverse KL divergence between the decoding and input language model's distribution.

\smallskip \noindent \textbf{Reasoning.} 
\citet{wang2023selfconsistency} propose \textsc{Self-Consistency}, which improves chain-of-thought prompting by replacing greedy decoding with a \textit{``sample-and-marginalize''} procedure: multiple diverse reasoning paths are sampled, then the most consistent answer is obtained by marginalizing over them. 
Additionally,
\citet{li2024escape} propose \textsc{ESC} (Early-stopping Self-Consistency), which improves efficiency by segmenting the large sample size into sequential small windows and halting sampling when all answers within a window converge, resulting in zero entropy in the predicted answer probability distribution.

\smallskip \noindent \textbf{Others.}
\citet{park2024grammaraligned} propose \textsc{ASAp} (Adaptive Sampling with Approximate Expected Futures), an adaptive sampling algorithm that ensures grammatical output while aligning with an LLM's probability distribution under grammar constraints. It iteratively refines grammaticality predictions by sampling and focusing on ungrammatical regions, ultimately converging to exact samples from the constrained probability distribution.

\subsection{Tree-Search-based Decoding}
\label{ssec:tree}

Planning algorithms, such as Monte-Carlo tree search (MCTS), have also been applied to identify optimal text outputs for various tasks.
For example,
\citet{zhang2023planning} propose \textsc{PG-TD} (Planning-Guided Transformer Decoding), a model-agnostic algorithm that enhances code generation by integrating an MCTS-inspired planning mechanism for lookahead search without requiring program grammar knowledge.
\citet{yu-etal-2023-prompt} propose \textsc{GDP-Zero} (Goal-oriented Dialogue Planning with Zero training) which prompts an LLM to simulate dialogue interactions at each stage of an Open-Loop MCTS. 
\citet{hao-etal-2023-reasoning} propose \textsc{RAP} (Reasoning via Planning (RAP), wherein the LLM interacts with a \textit{world model} -- the same LLM simulating the environment via specialized prompts -- and engages in planning via MCTS using candidate actions from the \textit{world model}.

\subsection{Model-level Decoding}
\label{ssec:model}

Model-level methods operate inside the intermediate layers of the model. 
\citet{gera2023autocontrastive} propose \textsc{ACD} (Autocontrastive Decoding), which leverages an \textit{early-exit} setup, contrasting the probability distributions from an early layer (anti-expert) and a later layer (expert).
\citet{zhang-etal-2024-draft} propose \textsc{Self-Speculative Decoding}, using a single LLM for drafting and verification. By selectively skipping intermediate layers with Bayesian optimization, draft tokens are generated efficiently and verified in single forward pass. 
To enhance factual accuracy,
\citet{zhang2024sled} propose \textsc{SLED} (Self Logits Evolution Decoding), which refines the logits with latent knowledge from early layers via minimizing KL divergence.
\citet{kojima-etal-2024-multilingual} propose \textsc{Language-Specific Neurons}, activating certain language-specific neurons during inference increases the probability of specific language occurrence (e.g. French) during generation. 
Similarly,
\citet{halawi2024overthinking} identify \textsc{Overthinking and False Induction Heads}, analyze where correct and incorrect classifications occur in terms of a \textit{critical layer}, then identify and remove the offending \textit{attention heads}, reducing the accuracy gap with minimal effects on correct prompts. 

\section{Context-Aware Self-Improvement}
\label{sec:contextawareselfimprovement}

Context-Aware Self-Improvement enhances performance using specialized prompt-based (\S \ref{ssec:prompting} and \S \ref{ssec:disturbed}) or retrieval-based (\S \ref{ssec:retrievalbased}) techniques. 

\subsection{Prompting}
\label{ssec:prompting}

Prompting uses carefully crafted prompts to enable few-shot or zero-shot learning without parameter updates~\cite{liu2023prompt}. 

\smallskip \noindent \textbf{Reasoning.}
\citet{wei2022chain} propose \textsc{CoT Prompting} (Chain-of-Thought Prompting), which allows language models to generate coherent reasoning steps leading to an answer, given few-shot examples of such reasoning.
Meanwhile,
\citet{kojima2022large} propose \textsc{Zero-Shot CoT Prompting}, which uses two-step template prompting: (i) \textit{``Let's think step by step''} to generate reasoning, followed by (ii) \textit{``Therefore, the answer is''} to extract the final answer.
To improve zero-shot CoT,
\citet{mekala-etal-2024-echoprompt} propose \textsc{EchoPrompt}, which adds query rephrasing as an initial step -- modifying the first prompt to \textit{``Let's repeat the question and also think step by step.''} 
However,
\citet{shaikh-etal-2023-second} evaluates zero-shot CoT in harmful questions and stereotype benchmarks, finding it \textit{significantly increases the likelihood of harmful outputs} across different prompts and models. 

\smallskip \noindent \textbf{Others.}
For methods used in other tasks,
\citet{wang2023decodingtrust} propose \textsc{DecodingTrust}, a multi-task (toxicity, stereotypes, adversarial robustness, out-of-distribution robustness, privacy, machine ethics, and fairness) \textit{trustworthiness evaluation suite} for LLMs with a focus on GPT models.
\citet{huang2024catastrophic} propose \textsc{Generation Exploitation Attack} which investigates system prompt strategies (prepending or excluding prompts) and decoding strategies, including temperature, Top-$K$, and Top-$p$ (nucleus) sampling. 
\citet{lin2024the} propose \textsc{URIAL} (Untuned LLMs with Restyled In-context ALignment), which aligns base LLMs using in-context learning with a few curated examples and a designed system prompt.

\subsection{Disturbed Prompt}
\label{ssec:disturbed}

Disturbed prompt methods use a specialized prompt (e.g., disturbance or noisy instruction) to obtain and contrast the model's probability distributions \textit{with and without the specialized prompt} during the decoding.
\citet{wang-etal-2024-mitigating} propose \textsc{ICD} (Instruction Contrastive Decoding), 
a method using disturbance instructions as role prefixes (e.g., \textit{``You are a confused object detector''}). The model's probability distributions \textit{with and without the disturbance instruction} are contrasted during decoding for hallucination reduction.
\citet{shi-etal-2024-trusting} propose \textsc{CAD} (Context-Aware Decoding), which contrasts the model's probability distributions \textit{with and without additional context} -- which may contain information that is contradictory to the model's prior knowledge -- during decoding for hallucination reduction.
\citet{kim2024instructive} propose \textsc{ID} (Instructive Decoding), where the model's probability distributions \textit{with and without the noisy instruction} are contrasted during decoding.
\citet{yuan-etal-2024-discerning} propose \textsc{COIECD} (Contextual Information-Entropy Constraint Decoding), an adaptive decoding strategy to mitigate knowledge conflicts between an LLM's parametric knowledge and contextual knowledge appended to the input text. The method applies an \textit{information-entropy constraint}, based on the \textit{stable entropy hypothesis} \cite{arora2023stable}, to identify and adjust \textit{non-compliant tokens} with relatively high or low entropy, improving generation by dampening the model's parametric knowledge.
For improving safety, \citet{zhong-etal-2024-rose} propose \textsc{ROSE} (Reverse prOmpt contraStive dEcoding), which contrasts outputs with \textit{a positive prompt and a negative prompt} during decoding.


\subsection{Retrieval-based}
\label{ssec:retrievalbased}

Retrieval-based methods obtain information from existing corpora or construct a retrieval datastore. 
\citet{Khandelwal2020Generalization} propose \textsc{$k$NN-LM}, which uses the hidden state as a query for token retrieval in a constructed key-value datastore, and then combines the probability distribution with the probability distribution of the aggregated retrieved neighbor similarity. 
To improve speculative decoding, 
\citet{li2024nearest} propose \textsc{NEST} (Nearest Neighbor Speculative Decoding), which extends \textsc{$k$NN-LM} by adding a passage retrieval step to minimize token storage and search.
Similarly, \citet{he-etal-2024-rest} propose \textsc{REST} (Retrieval-Based Speculative Decoding), which replaces the draft model with a retrieval datastore for speculative decoding, using previous tokens to find exact matches and organizing candidates in a Trie to select the most frequent nodes as draft tokens.
To improve decoding trustworthiness, 
\citet{luohe2024reference} propose \textsc{RTD} (Reference Trustable Decoding), a method that retrieves the top-$k$ references from a constructed reference datastore, normalizes and aggregates them into a reference probability distribution, and combines them with the original LLM distribution.
\cite{zhao-etal-2024-enhancing} propose \textsc{Multi-Input CD} (Multi-Input Contrastive Decoding), where the model combines the logits of its parametric knowledge with contrastive predictions from both relevant contexts (top retrieved texts from an external knowledge base) and irrelevant contexts (adversarially crafted or low-ranked texts).

\section{Model-Aided Self-Improvement}
\label{sec:modelaided}

Model-Aided Self-Improvement enhances performance with an external (often small) model, such as an (Anti-)Expert model (\S \ref{sec:expert-anti-expert}), draft model (\S \ref{sec:draft-model}), small amateur model (\S \ref{ssec:smalllm}), reward model (\S \ref{sec:reward-model}) and tool/APIs (\S \ref{sec:tool-use}).

\subsection{Expert and/or Anti-Expert} 
\label{sec:expert-anti-expert}

Expert and/or Anti-Expert models -- \textit{specialized or not} in a particular task -- provide logits or probability distributions, which are then contrasted or incorporated during decoding, or for otherwise guiding the decoding process via scoring.


\smallskip \noindent \textbf{Toxicity.}
\citet{liu-etal-2021-dexperts} propose \textsc{DExperts} (Decoding-Time Experts), wherein they combine a neutral model, expert model, and anti-expert model via their logits.
\citet{zhang-wan-2023-mil} propose \textsc{MIL-Decoding} (Multiple Instance Learning Decoding) via an MIL network which assigns toxicity probabilities to tokens, and then combines them with the LLM probability distribution to form detoxified text.

\smallskip \noindent \textbf{Machine Translation.}
\citet{wang-etal-2023-easy} propose \textsc{PSGD} (Prefix-Suffix Guided Decoding) -- for machine translation suggestion. By using special prefix and suffix tokens and modifying the decoding objective, the model only generates for marked spans.
\citet{zeng-etal-2024-improving} propose \textsc{CoDec} (Cooperative Decoding), which treats an initial translation from a traditional Neural Machine Translation model as a draft, and then uses another LLM-based Machine Translation model to verify and correct (via decoding that segment itself) the translation.
\citet{yang-etal-2024-language-informed} propose \textsc{LiBS} (Language-informed Beam Search), a method for ensuring machine translation occurs in the correct output language. The method incorporates a language identification model into the beam search process via ranking the candidate beams.
\citet{le2024constrained} propose \textsc{CODEC} (Constraint Decoding for Cross-lingual Label Projection), which translates high-resource data. First, the translation occurs \textit{without markers}, then \textit{with markers} on a second decoding pass. The two translations are then compared via their log-probability distributions. Then, the top-$k$ hypotheses are re-ranked, and decoding time is optimized with a depth-first search and branch-and-bound approach.

\smallskip \noindent \textbf{Alignment.}
To improve the model alignment,
\citet{shi2024decodingtime} propose \textsc{MOD} (Multi-Objective Decoding), which merges the predictive distributions of $n$ models trained for individual objectives (e.g. \textit{helpfulness}, \textit{safety}) to align the $n$ language models for multi-task settings via a closed-form solution inspired by the Legendre transform.
\citet{chakraborty2024transfer} propose \textsc{Transfer $Q^*$}, an indirect transfer approach that computes the optimal value function for a target reward by leveraging a baseline \textit{aligned model}, which is aligned with a potentially different baseline reward.

\smallskip \noindent \textbf{Others.}
For other tasks,
\citet{xu-etal-2024-safedecoding} propose \textsc{SafeDecoding}, a safety-aware decoding strategy, which mitigates jailbreak attacks by constructing a probability distribution from both the original LLM and a safety expert model.
\citet{huang2023grounded} propose \textsc{GD} (Grounded Decoding), which solves robotic tasks by combining the probability distributions of an LLM and multiple \textit{Grounded Models}, and then selecting high-probability tokens from this combined distribution. 
\citet{shen2024superposed} propose \textsc{Superposed Decoding}, which generates $k$ drafts in one autoregressive pass. It combines them with the top-$k$ tokens to create $k^2$ options, caching the $k$ most likely tokens and using \textit{$n$-gram models} to filter incoherent generations.
\citet{jacob2024the} propose \textsc{Equilibrium-Ranking}, a game-theoretic method for consensus planning that combines generative and discriminative decoding by modeling it as a signaling game and adjusting only signaling policies to approximate equilibria.

\subsection{Draft Model}
\label{sec:draft-model}

We focus on the draft model used in \textsc{Speculative Decoding}~\cite{pmlr-v202-leviathan23a}, a method which accelerates LLM inference. The smaller draft model proposes multiple completions, and then leverages the target LLM to verify them in parallel. These off-the-shelf small LMs require no additional training or modifications to their architectures, enabling rapid adoption.

\citet{sun2023spectr} propose \textsc{SpecTr}, which uses an \textit{optimal transport-based} draft selection algorithm to achieve faster draft verification.
To improve the acceptance rate of draft tokens, \citet{gong-etal-2024-graph} propose \textsc{GSD} (Graph-structured Speculative Decoding), which uses a directed acyclic graph to generate multiple hypotheses, giving the model more options to select the longest valid sequence.
\citet{svirschevski2024specexec} propose \textsc{SpecExec} (Speculative Execution), which constructs a large draft ``cache'' tree of likely prefix continuations using parallel search, which is then validated in a single pass using the target model.
Similarly, \citet{chen2024sequoia} propose \textsc{SEQUOIA}, which leverages dynamic programming to create optimal tree structures for speculated tokens. This method enhances scalability and refines SpecInfer~\cite{miao2024specinfer} with replacement-free sampling from the draft model to boost robustness.
\citet{yin2024a} provide a theoretical exploration of Speculative Decoding through Markov chain abstraction.
\citet{yuan-etal-2024-speculative} propose \textsc{SCD} (Speculative Contrastive Decoding), which integrates the contrastive distribution to validate token accuracy.

Several works focus on refining the draft model itself for increased performance.
\citet{liu2024online} propose \textsc{Online Speculative Decoding}, which dynamically updates draft models based on user query data. This adaptation reduces distribution mismatches, improving the draft model's predictions,
Complementarily,
\citet{du2024glide} propose two improvements to vanilla speculative decoding to enhance decoding speed: \textsc{GliDe} (Glimpse Draft Model) and \textsc{CaPE} (Confidence-aware Proposal Expansion). \textsc{GliDe} improves the draft model by reusing cached keys and values from the target LLM, while \textsc{CaPE} leverages confidence scores from the draft model to verify additional candidate tokens.\footnote{See~\citet{xia-etal-2024-unlocking} for more on speculative decoding.}

\subsection{Small LM/Amateur LM}
\label{ssec:smalllm}

In addition to the Draft Model used in Speculative Decoding, other Small LMs -- also referred to as Amateur LMs --  are used to guide LLM generation. 

\smallskip \noindent \textbf{Classification Model.}
Various types of classification models are employed to guide generation, such as those used for evaluation, probability adjustment, and token replacement.
\citet{miyano-etal-2023-self} propose \textsc{NeuroLogic-A* (P)} (a version of NeuroLogic-A* with positive constraints). This method joins high-quality fragments from the N-best hypothesis generations via a token-level binary classification model (for token quality estimation) which is used to formulate \textit{constraints} for input to NeuroLogic-A*.
\citet{lango-dusek-2023-critic} propose \textsc{Critic-Driven Decoding}, via a \textit{text critic classifier}. The output of the classifier is combined with the LM probability distribution via a scaling factor.
Similarly,
\citet{choi-etal-2023-kcts} propose \textsc{KCTS} (Knowledge-Constrained Tree Search), which steers LLM-generated outputs to align with reference knowledge by integrating a knowledge classifier score with Monte-Carlo Tree Search (MCTS). The candidate generation is formulated as the root of the tree, and other generations are sampled and incorporated via leaf nodes, where the factuality is evaluated via a token-level knowledge groundedness metric.
\citet{zhang-etal-2024-jailbreak} propose \textsc{EnDec} (Enforced Decoding), a method to attack LLMs with a conditionally-activated decoding objective, which activates: (i) \textit{affirmative prefixes}, i.e., the insertion of a positive prefix (``Sure, here is''), and (ii) \textit{negation reversing}, i.e., the replacement of negative words with positive words (via sentiment classification).

\smallskip \noindent \textbf{Generative Model.}
\citet{li-etal-2023-contrastive} propose \textsc{CD} (Contrastive Decoding), which contrasts the probability distributions of a large expert model and a small amateur model, penalizing next-tokens where the amateur model is highly confident. This decoding objective is active only when the expert language model does not obtain high confidence scores, via an \textit{adaptive plausibility constraint} also incorporated into the decoding objective.
\citet{kim2023speculative} propose \textsc{BiLD} (Big Little Decoder), a framework that learns a policy to use two models of different sizes to generate text through fallback and rollback strategies collaboratively. The small model generates text autoregressively with low inference cost, while the large model occasionally refines its errors non-autoregressively.
\citet{fisher-etal-2024-jamdec} propose \textsc{JAMDEC} (JAMbalaya DECoding), an authorship obfuscation method using small language models. It involves (i) keyword extraction, (ii) over-generation via \textit{Constrained Diverse Beam Search}, which integrates lexical constraints -- the keywords -- and a diversity with a similarity-based penalty into the decoding objective, and (iii) filtering by quality and content overlap thresholds to select the most stylistically distinct candidate.

\subsection{Reward Model}
\label{sec:reward-model}

The reward model is a fine-tuned LM that evaluates generated responses and assigns scores to guide the decoding process. \citet{deng-raffel-2023-reward} propose \textsc{RAD} (Reward-Augmented Decoding). This method uses an external reward model to reweight the logits with a \textit{steering factor} hyperparameter to control the reweighting strength. 
To improve the alignment, \citet{khanov2024args} propose \textsc{ARGS} (Alignment as Reward-Guided Search), which integrates alignment into the decoding process. A reward mechanism is used to adjust the model's probability distribution at each step, and token selection happens via either greedy or stochastic sampling at each step.
\citet{wan2024alphazerolike} propose \textsc{TS-LLM} (Tree-Search enhanced LLM), which extends prior work by integrating an AlphaZero-like deep \textit{tree-search} with a learned LLM-based value function, conditioned on state and a learned outcome reward model.
In addition, multiple scorers can be combined at inference to tackle multi-objective RL without extra training: \citet{mudgal2024controlled} propose \textsc{CD} (Controlled Decoding), which uses a trained prefix scorer to guide generation from a fixed model, solving KL-regularized RL objectives.

\subsection{Tool Use/APIs}
\label{sec:tool-use}

Interaction with external models also includes tool use -- such as parsers, static analysis tools, and API calls.
For example,
\citet{geng-etal-2023-grammar} propose \textsc{GCD} (Grammar Constrained Decoding), which uses specific context-free grammars for various highly-structured tasks to enforce constraints via an incremental parser, pruning non-compliant tokens -- \textit{forbidden tokens} -- from the probability distribution at each decoding timestep, and keeping only \textit{allowed tokens} with respect to the task-specific formal grammar. 
\citet{bastan-etal-2023-neurostructural} propose \textsc{NeuroStructural Decoding}, which applies structural lexical and syntactic constraints to preserve subject-verb-object relationships.
It extends \textsc{NeuroLogic Decoding} \cite{lu-etal-2021-neurologic} by incorporating constraints in conjunctive normal form, and scoring candidate generations -- modifying the beam search process -- considering the probability distribution and the quantity of clauses satisfied. 
In code generation, \citet{agrawal2023monitorguided} propose \textsc{MGD} (Monitor-Guided Decoding), wherein a \textit{monitor} is integrated into the LM decoding process, which queries a static analysis tool; the output is then incorporated via a mask on the logits. 
\citet{roy-etal-2024-flap} propose \textsc{FLAP} (Flow-Adhering Planning) for dialogue systems. The planning steps (\textit{``flow''}) and corresponding API calls are converted into two dependency graphs, which are used as constraints during decoding with a lookahead heuristic added to the probability distribution.

\section{Discussion}\label{sec:discussion}

Inference-time self-improvement methods excel in reasoning (\S \ref{ssec:sampling}, \S \ref{ssec:prompting}), 
enable faithful generation (\S \ref{ssec:contrastive}, \S \ref{ssec:retrievalbased}), 
increase speed via parallelism (\S \ref{ssec:parallel}), 
and more without updating model parameters or additional training. 
Despite these advancements, several challenges remain. In this section, we discuss considerations for method selection, and outline potential directions for future research:

\smallskip \noindent
\textbf{Maintenance}: Methods with dependence on an external datastore (\S \ref{sec:contextawareselfimprovement}) or model (\S \ref{sec:modelaided}) require ongoing maintenance, as they need to be updated over time. In contrast, independent methods (\S \ref{sec:self}) do not require this level of maintenance, as they solely operate based on the decoding process.

\smallskip \noindent
\textbf{Amount of Model Access}: For example, the OpenAI Chat API returns a partial probability distribution -- i.e. \textit{not} over the entire token vocabulary -- therefore methods which operate on the full probability distribution are not a suitable option.

\smallskip \noindent
\textbf{Trade-Offs in Inference Costs}: Methods scaling up inference time such as sampling with multiple generations (\S \ref{ssec:sampling}) generally take more time at inference than methods directly manipulating the decoding process.

\smallskip \noindent
\textbf{Generalizability}: External models used to guide the generation process are typically specialized for specific domains or tasks. Adapting these models to new contexts or unseen data often necessitates creating new expert models or additional fine-tuning. Moreover, the overall performance is closely tied to the quality of auxiliary models.

\smallskip \noindent
\textbf{Generation Quality}:
Methods manipulating the decoding process offer significant flexibility, enabling real-time adjustments to the generation process to satisfy specific control conditions (\S \ref{ssec:constrained}). However, these control conditions are competing constraints with the LLM's inherent language generation tendencies towards fluency and coherence. Hence, there can be a trade-off between enforcing constraints and maintaining generation quality.

\smallskip \noindent
\textbf{Explainability and Interpretability}:
Only a few works~\cite{kojima-etal-2024-multilingual,halawi2024overthinking} analyze the LLM decoding process from the perspective of neurons and attention heads, or provide a theoretical analysis~\cite{yin2024a} of the decoding process. This is an opportunity for future work to develop more robust methods for understanding the decision-making processes of LLMs~\cite{bismay2024reasoningrec,singh2024rethinking}, particularly for complex tasks like reasoning~\cite{wei-jie-etal-2024-interpretable}.

\section*{Limitations}

Due to space limitations, we cannot present all methods with exhaustive technical details. Our focus is primarily on methods from key sources such as ACL, EMNLP, NAACL, NeurIPS, ICLR, ICML, and arXiv in recent years. We plan to continuously follow these sources and incorporate new methods and datasets in our online resource (see \S \ref{sec:introduction}).


\section*{Ethical Considerations}
This paper reviews the challenges, methods, and limitations of LLM self-improvement in the existing literature, bringing up some important ethical issues. 

\paragraph{Social Bias.} 
LLMs have long been known to exhibit bias with respect to social groups -- i.e. based on race, gender, and more \cite{caliskan2017semantics, may-etal-2019-measuring, blodgett-etal-2020-language, meade-etal-2022-empirical, teleki25_icwsm}. Similarly to the need for bias analysis and corresponding debiasing methods for (i) base LLMs \cite{sun-etal-2019-mitigating, zmigrod-etal-2019-counterfactual, dong2024disclosure}, and (ii) parameter-efficient methods \cite{xie-lukasiewicz-2023-empirical, dong-etal-2023-co2pt}, bias analysis and corresponding debiasing methods will be needed for self-improvement methods. For example, \citet{shaikh-etal-2023-second} find that Zero-shot CoT significantly increases the likelihood of harmful outputs. An effective strategy for reducing these biases may be including explicit mitigation instructions in the prompt~\cite{si2023prompting}. An additional consideration is that methods which incorporate external data or models may introduce biases in this way as well.

\paragraph{Economic Equity.} LLMs are cost-prohibitive -- consider the colloquial expression: \textit{``GPU poor.''} Lightweight, inference-time methods such as via self-improvement have the potential to decrease the cost of LLM access in Model-as-a-Service settings and in locally-hosted settings, potentially increasing economic equity.

\paragraph{Environmental Sustainability.} Efficient methods beget environmental benefits,$^1$ decreasing the total carbon footprint of LLM-associated activities.

\bibliography{arXiv}

\clearpage
\appendix

\begin{table*}[ht]
\centering
\caption{Tasks with representative methods.}
\label{tab:tasks}
\resizebox{\linewidth}{!}{%
\begin{tabular}{lp{15cm}} \hline
\textbf{Tasks}                     & \textbf{Representative Method(s)} \\ \hline 
Jailbreak and Defense              & \cite{huang2024catastrophic, zhang-etal-2024-jailbreak, xu-etal-2024-safedecoding} \\
Stereotype and Toxicity Benchmarks & \cite{shaikh-etal-2023-second, wang2023decodingtrust} \\
Code Generation                    & \cite{zhang-etal-2024-draft, zhang2023planning} \\
Dialogue                           & \cite{yao-etal-2023-fine, nandwani-etal-2023-pointwise, roy-etal-2024-flap, he-etal-2024-rest, chen-etal-2024-control} \\
Open-Ended Generation              & \cite{li-etal-2023-contrastive, ji2024language, zhang2024sled, zhu2024improving} \\
Question-Answering                 & \cite{yuan-etal-2024-discerning, chuang2024dola, zhao-etal-2024-enhancing, jacob2024the} \\
Alignment                          & \cite{khanov2024args, lin2024the, shi2024decodingtime, chakraborty2024transfer} \\
Hallucinations and factuality      & \cite{manevich-tsarfaty-2024-mitigating, wang-etal-2024-mitigating, chuang2024dola} \\
Reasoning                          & \cite{wang2023selfconsistency, mekala-etal-2024-echoprompt, li2024escape, wei2022chain, wang2023selfconsistency, wan2024alphazerolike, hao-etal-2023-reasoning} \\
Machine Translation                & \cite{deguchi-etal-2024-centroid, zeng-etal-2024-improving, yang-etal-2024-language-informed, sia-etal-2024-anti, pmlr-v202-vilnis23a, trabelsi2024efficient} \\ \hline                                                                    
\end{tabular}%
}
\end{table*}

\section*{Appendix}

\section{Definitions}
\label{sec:definitions}

We provide a brief overview of related definitions,\footnote{For a more detailed overview of fundamental definitions, we refer readers to \citet{ippolito-etal-2019-comparison}.} following \citet{ippolito-etal-2019-comparison, sia-etal-2024-anti, yuan-etal-2024-discerning}. We assume an autoregressive formulation (i.e., depending on $y_{<t}$, the previously generated tokens at each timestep, $t$). 

\textbf{Decoding Objective} refers to the algorithm for next-token selection via selecting the token with the maximum log-likelihood from the next-token probability distribution. We present \textit{deterministic greedy decoding}:

\begin{equation*} \label{eq:decoding-objective}
y_t = 
\underset{T}{\mathrm{argmax}}\; 
log \space \mathcal{P}(y_t|y_{<t},x,p_{u},p_{sys})
\end{equation*}

\noindent where $y_t$ is the selected next-token, $\mathcal{P}(y_t|y_{<t},x,p_{u},p_{sys})$ is the probability distribution over all tokens in the vocabulary, $V$ of size $|V|$, conditioned on $y_{<t}$, the previously generated tokens, $x$, the input, $p_{u}$, the user prompt, and $p_{sys}$, the system prompt. In \textit{beam search decoding}, the top $k$ tokens are selected in a process which forms multiple generation {hypotheses}, and eventually the highest-scoring hypothesis (based on log-likelihood) is kept. In \textit{stochastic methods}, the next token is selected with probability $\mathcal{P}$, which can be reshaped using a temperature hyperparameter, $\tau$, to control the uniformity of $\mathcal{P}$: $\tau>1$ flattens the distribution, meaning that the next token selection will be more uniform, vice-versa $\tau<1$ sharpens the distribution, and $\tau=1$ corresponds to the original, unmodified distribution $\mathcal{P}$.



\textbf{Probability Distribution} or \textbf{Likelihood} refers to the next token  $w_i \in V$, distribution, obtained via the \textit{softmax} normalization function -- i.e., $z_i \rightarrow [0,1]$ for token vocabulary, $V$:

\begin{equation*} \label{eq:prob-distb}
\mathcal{P}(y_{t}=w_i|y_{<t},x,p_{u},p_{sys}) = 
\frac
{e^{z_{t,i}}}
{\sum_{j=1}^{|V|} e^{z_{t,j}}}
\end{equation*}


\textbf{Logits} refers to the raw outputs of the model, $\mathcal{F}$, before the \textit{softmax} function is applied:

\begin{equation*} \label{eq:next-token-logits}
z_t=\mathcal{F}(y_{<t},x,p_{u},p_{sys})
\end{equation*}

\textbf{Entropy} refers to the overall confidence of the language model in predicting the next token $y_t$ given the already generated tokens $y_{<t}$ and the input $x$ (we omit $p_{u}$ and $p_{sys}$ for simplicity):

\begin{equation*} \label{eq:entropy}
\mathcal{H}
(y_t|y_{<t},x)
=
\underset{y_t \sim \mathcal{P}(\cdot | y_{<t})}{\mathbb{E}}\;
-\log 
\mathcal{P}
(y_t|y_{<t},x) 
\end{equation*}

Where $\mathbb{E}$ is the \textit{expectation} of $\mathcal{P}$ -- i.e., a weighted average of all possible outcomes. Hence, for high values of $\mathbb{E}$, the interpretation is that the model has high confidence in the generation (indicated by the overall high values of $\mathcal{P}$), and vice versa for low values of $\mathbb{E}$.

\section{Comparison with PEFT Methods}

Parameter-efficient fine-tuning (PEFT) methods include LoRA \cite{hu2022lora}, ProLoRA \cite{wang-etal-2024-prolora}, DoRA \cite{liu2024dora}, and others. These methods involve using a low-rank matrix to approximate the model weights for fine-tuning purposes, hence reducing computational complexity.

\paragraph{Need for Dataset.} PEFT methods are still \textit{fine-tuning methods}, meaning they still rely on datasets. There are situations where it may be easier to access a model (\S \ref{sec:modelaided}) than a dataset -- hence inference-time self-improvement methods are a better fit in many situations.

\paragraph{Need for Model Parameters.} PEFT methods require access to the original model parameters -- however, many inference-time self-improvement methods do \textit{not} require this type of access. Hence, inference-time self-improvement methods are a better fit in many situations.

\section{Applications}

We present some tasks with representative methods mentioned in this paper in Table \ref{tab:tasks}.

\end{document}